\pdfoutput=1

\documentclass[11pt]{article}

\usepackage[preprint]{acl}

\usepackage{times}
\usepackage{latexsym}

\usepackage[T1]{fontenc}

\usepackage[utf8]{inputenc}

\usepackage{microtype}

\usepackage{inconsolata}

\usepackage{graphicx}

\usepackage{amsmath}
\usepackage{amssymb}
\usepackage{booktabs}
\usepackage{multirow}
\usepackage{xcolor,colortbl}
\usepackage{graphicx}
\usepackage{subcaption}
\usepackage{pifont}
\newcommand{\cmark}{\ding{51}}
\newcommand{\xmark}{\ding{55}}
\usepackage{enumitem}

\renewcommand{\paragraph}[1]{{\vspace{0.05in}\noindent\textbf{#1}}\quad}

\newcommand{\eg}{\textit{e.g.}}
\newcommand{\ie}{\textit{i.e.}}
\definecolor{Gray}{gray}{0.9}
\definecolor{LightCyan}{rgb}{0.92,1,1}
\definecolor{color_ours}{gray}{0.9}

\newcolumntype{a}{>{\columncolor{color_ours}}c}

\title{Joint Fusion and Encoding: \\
Advancing Multimodal Retrieval from the Ground Up}

\author{Lang Huang$^{*}$\, Qiyu Wu\thanks{Equal contribution.~ $^\dag$Work done in a personal capacity.}$^\dag$, 
        Zhongtong Miao, Toshihiko Yamasaki \\
       \normalfont{The University of Tokyo, Tokyo, Japan}
}
\newcommand{\modelname}{JFE}
\newcommand{\modelfullname}{Joint Fusion Encoder}

\begin{document}
\maketitle
\footnotetext[2]{Correspondence to: \{langhuang, yamasaki\}@cvm.t.u-tokyo.ac.jp and wuqiyu576@gmail.com}

\begin{abstract}

Information retrieval is indispensable for today’s Internet applications, yet traditional semantic matching techniques often fall short in capturing the fine-grained cross-modal interactions required for complex queries. Although late-fusion two-tower architectures attempt to bridge this gap by independently encoding visual and textual data before merging them at a high level, they frequently overlook the subtle interplay essential for comprehensive understanding.
In this work, we rigorously assess these limitations and introduce a unified retrieval framework that fuses visual and textual cues from the ground up, enabling early cross-modal interactions for enhancing context interpretation.
Through a two-stage training process—comprising post-training adaptation followed by instruction tuning—we adapt MLLMs as retrievers using a simple one-tower architecture. Our approach outperforms conventional methods across diverse retrieval scenarios, particularly when processing complex multi-modal inputs. Notably, the joint fusion encoder yields greater improvements on tasks that require modality fusion compared to those that do not, underscoring the transformative potential of early integration strategies and pointing toward a promising direction for contextually aware and effective information retrieval.

\end{abstract}
\begin{figure}[t]
    \centering
    \includegraphics[width=\linewidth]{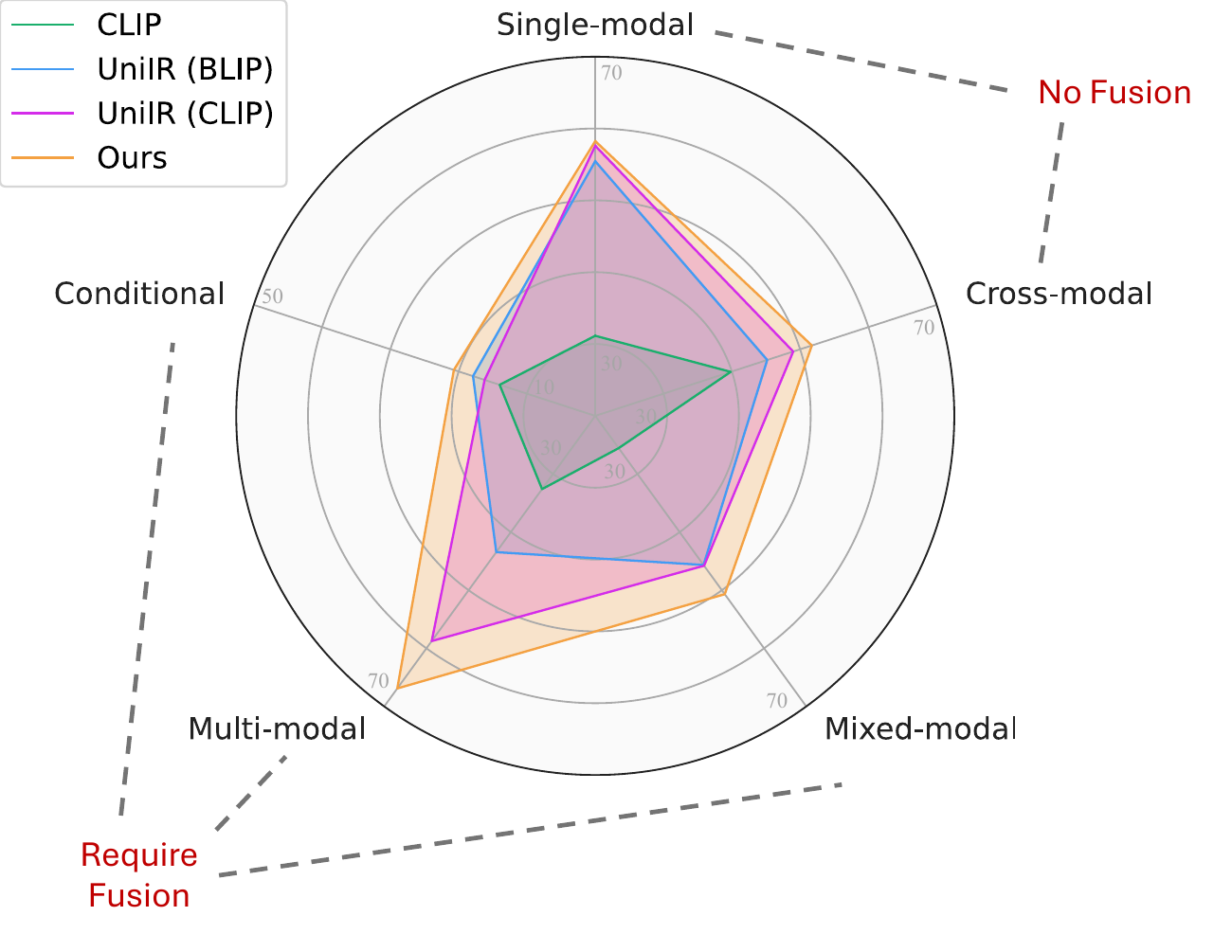}
    \vspace{-0.15in}
    \caption{Multi-modal retrieval results on a collection of retrieval tasks, categorized as 1) Single-modal, 2) Cross-modal, 3) Mixed-modal, 4) Multi-modal, and 5) Conditional. CLIP~\cite{radford2021learning} is two-tower model, UniIR~\cite{wei2024uniir} is two-legs model with late-fusion, and ours is built as one-tower with early fusion. Our approach obtains moderate improvements in retrieval tasks requiring no modality fusion compared to state-of-the-art two-tower methods, the gains become prominent when the tasks need modality fusion that involve multiple modalities or conditional information in the query or candidate. Specific results are in \S\ref{sec:exp}.
    }
    \label{fig:teaser}
    \vspace{-0.1in}
\end{figure}

\begin{figure*}[t]
    \centering
    \includegraphics[width=\textwidth]{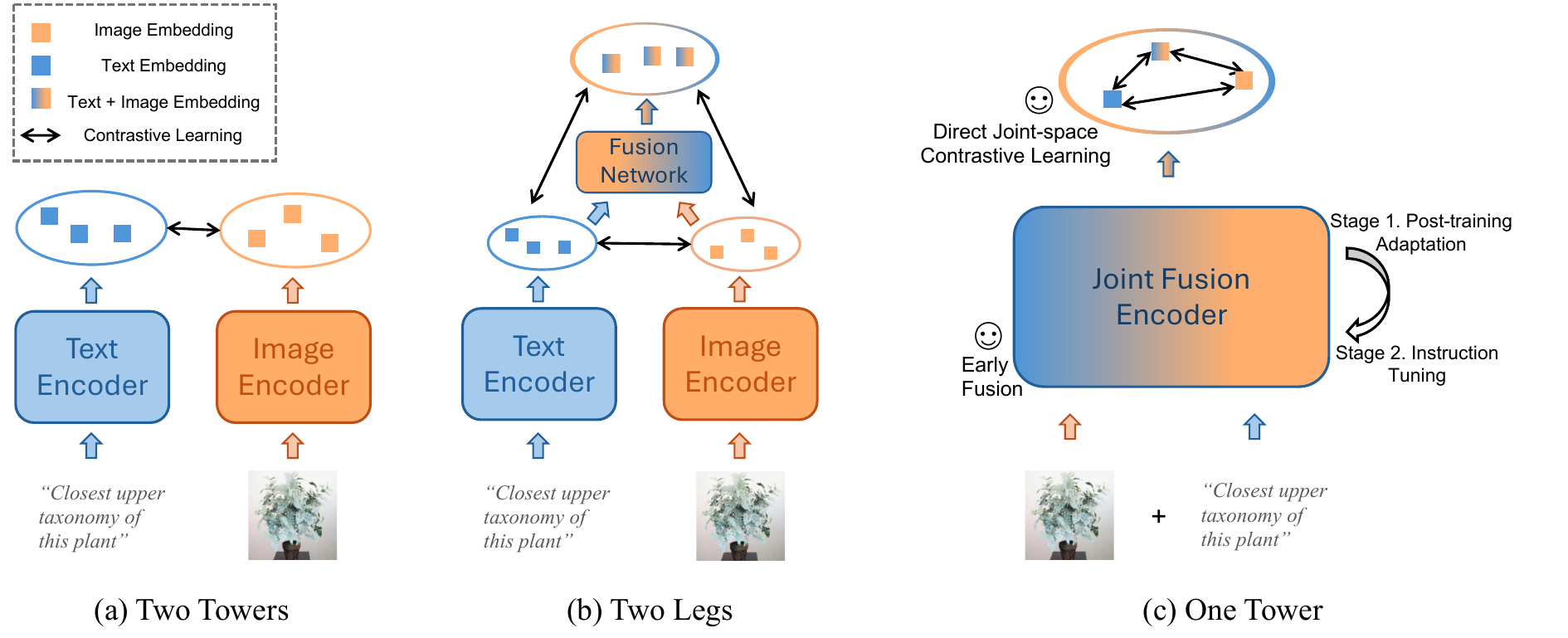}
    \caption{Conceptual illustration of multi-modal retrieval frameworks.
        \textit{(a)} Traditional two-tower methods encode text and images separately, limiting early cross-modal interactions and often overlooking nuanced user queries.
        \textit{(b)} Two-leg methods introduce an additional fusion network to merge single-modal embeddings, enabling more complex tasks yet still deferring cross-modal interplay until later stages.
        \textit{(c)} Our proposed \modelfullname{}~(\modelname{}), an one-tower method, integrates visual and textual cues from the ground up, unifying the embedding space for a direct contrastive learning and facilitating fine-grained multi-modal understanding. This approach, including two stages: post-training adaptation and instruction tuning, simplifies and improves the retriever when complex multi-modal understanding is required.}
    \label{fig:vlretrieval}
\end{figure*}

\section{Introduction}
\label{sec:intro}
Retrieval is a cornerstone task in modern Internet systems and artificial intelligence, traditionally built on semantic matching to identify relevant information from vast datasets. Early retrieval methods~\cite{robertson1995okapi, gao-etal-2021-simcse, wu-etal-2022-pcl} focused exclusively on single modalities--primarily text--which yielded effective systems in limited settings but inherently lacked the capacity to process multi-modal inputs. As applications evolved to require richer, cross-modal information, such as integrating visual content alongside text, the limitations of these conventional methods became increasingly evident. In response, researchers adapted traditional approaches~\cite{radford2021learning} using late-fusion~\cite{wei2024uniir}, two-tower architectures, where each modality is encoded separately and only fused at a high level. Although this strategy bridges the gap between modalities, it falls short in capturing the nuanced, early interactions, and fine-grained details essential for fully understanding complex user intentions. In emerging applications such as retrieval-augmented generation~\cite{lewis2020retrieval,gao2023retrieval}, context-aware search~\cite{han2017fashion200k, Liu_2021_ICCV}, and conditional search~\cite{Vaze2023GeneCIS}, the simple high-level fusion of independent representations is insufficient. Instead, these scenarios demand an integrated approach that can blend visual and textual cues from the ground up, thereby enabling more precise interpretation of intricate, multi-modal queries.

In this work, we propose a unified multi-modal retrieval framework, \modelfullname{}~(\modelname{}), that seamlessly interweaves encoding and fusion into a single process, fundamentally rethinking the way retrieval systems handle complex, heterogeneous queries. As shown in Figure~\ref{fig:vlretrieval}, conventional approaches retrofit single-modal systems to accommodate multi-modal tasks by independently encoding each modality and subsequently merging their representations--a late-fusion strategy that often fails to capture the nuanced interplay between textual and visual cues. In contrast, one-tower architecture embraces a joint fusion and encoding paradigm, wherein fusion occurs concurrently with encoding. This early integration addresses the intrinsic challenge of harmonizing the disparate characteristics of multi-modal data, enabling the capture of fine-grained, inter-modal interactions and preserving the modality-specific nuances critical for deciphering intricate user intentions.

Besides above fundamental benefits~\cite{jang2023unifying, li2024multimodal} of such one-tower paradigm for multi-modal representation, recent advent of large-scale pre-trained large language models (LLMs) and multi-modal LLMs (MLLMs) built on them, has made the one-tower design become even more compelling.
By repurposing MLLMs and adapting them into effective encoders through contrastive learning and instruction-tuning, our approach not only simplifies the retrieval pipeline by eliminating the need for separate fusion modules but also produces unified, powerful embeddings that significantly enhance performance across a range of multi-modal retrieval tasks. The proposed approach consists of two key stages:
1)  Post-training adaptation:
We begin by adapting the MLLM, which was originally pre-trained as an autoregressive decoder, to function as an encoder. This is achieved through post-training using contrastive learning loss, allowing the model to effectively process inputs into meaningful embeddings;
2) Instruction tuning
We then fine-tune the model with instructive information specific to various cross-modal retrieval tasks. This step enables the model to better understand and follow instructions in both textual and visual contexts.

To thoroughly evaluate the capabilities of multi-modal retrieval models, we have organized extensive experiments across a comprehensive set of benchmarks, categorized by their input and output modalities: 1) single-modal, both the query and candidate are in the same modality; 2) cross-modal, the query and candidate are in different modalities; 3) mixed-modal, either query and candidate is multi-modal and the other remains single-modal; 4) multi-modal, both query and candidate are multi-modal; and 5) conditional, the query is supplemented with additional contextual or conditional information.
As shown in Figure~\ref{fig:teaser}, we can observe that while our joint fusion and encoding method \modelname{} obtains moderate improvements in standard single-modal and cross-modal retrieval compared to state-of-the-art two-tower methods, the gains become even prominent when the query and/or candidate involve multiple modalities or conditional information, the scenarios attaining mounting attention as massive multi-modal contents and instructions are produced every second.

\section{Related Work}










Multi-modal retrieval serves as a cornerstone for multi-modal information systems.
Exisiting studies in this field have been mostly based on two-towers or two-legs architectures~\citep{radford2021learning,9879245,liu2023universal,koukounas2024jina} as we shown in Figure~\ref{fig:vlretrieval}.
A notable example is CLIP~\citep{radford2021learning}, which has been widely adopted as a foundational representation model in multi-modal tasks, including retrieval. CLIP employs separate encoders for text and images and aligns them within a shared embedding space. Building on CLIP, Pic2Word~\citep{Saito2023Pic2Word} leverages pseudo language tokens to train a mapping network for zero-shot composed image retrieval, while SEARLE~\citep{Baldrati2023CIRCO} adopts a similar strategy by pre-training a textual inversion network for zero-shot composed image retrieval. 
Additionally, UniIR~\citep{wei2024uniir} utilizes score-level fusion and feature-level fusion as a comprehensive exploration of two-legs architectures, and improve CLIP/BLIP-based multi-modal information retrieval systems.

Meanwhile, there are latest works applying one-tower visual LLM, given the strong multi-modal understanding ability from large-scale pre-training. ~\citet{jiang2024vlm2vec} and \citet{zhang2024gme} employ visual-language models as its backbone to fuse textual and visual spaces into a unified vision-language embedding model, and they also introduces a comprehensive embedding dataset for evaluation on such a complex multimodal scenairo. Meanwhile, ~\citet{lin2024mm} adopts a similar approach using a visual-language model for retrieval, but it focuses on hard-negative sampling and leverages an LLM for reranking. Both of these concurrent works leverage visual-language models to fuse text and image inputs—thereby adopting a one-tower architecture similar to ours. They implicitly share our advocation that a one-tower architecture is particularly effective for complex vision-language representation and retrieval tasks.

\newcommand{\tss}[2]{{#1}^{\text{#2}}} 
\newcommand{\tsbs}[2]{{#1}_{\text{#2}}} 
\newcommand{\tssbs}[3]{{#1}_{\text{#2}}^{\text{#3}}} 

\section{Methodology}
\subsection{Multi-modal retrieval}
\label{sec:pre}
This section describes multi-modal retrieval, a task designed to retrieve relevant information from multi-modal databases by matching queries and candidates across textual, visual, or combined modalities. Formally, the query set $\mathcal{Q}$ and candidate set $\mathcal{C}$ can be defined as follows,
\begin{equation}    
\begin{aligned}
    \mathcal{Q} &= \{q \mid q \in \{\tss{q}{i}, \tss{q}{t}, \{\tss{q}{i}, \tss{q}{t}\}\}\} \\
    \mathcal{C} &= \{c \mid q \in \{\tss{c}{i}, \tss{c}{t}, \{\tss{c}{i}, \tss{c}{t}\}\}\}. \\
\end{aligned}
\label{eq:qc}
\end{equation}
Depending on application scenarios as introduced in \S\ref{sec:intro}, the query can be a text query $\tss{q}{t}$, an image query $\tss{q}{i}$ or a pair of a text and an image, \ie, $\{\tss{q}{i}, \tss{q}{t}\}$. Similarly, $\tss{c}{t}$, $\tss{c}{i}$, $\{\tss{c}{i}, \tss{c}{t}\}$ represent text candidate, image candidate, and Multi-modal candidate, respectively.

\paragraph{Conventional two-tower models.}
A conventional two-tower model, \eg, CLIP~\cite{radford2021learning}, encodes the inputs into meaningful embeddings using encoders for texts and images separately, as follows,
\begin{equation}
\begin{aligned}
    \tss{\mathbf{h}}{q-i} &= f_{\tss{\theta}{i}}(\tss{q}{i}) \in \mathbb{R}^D \\
    \tss{\mathbf{h}}{q-t} &= f_{\tss{\theta}{t}}(\tss{q}{t}) \in \mathbb{R}^D
\end{aligned}
\label{eq:bienc}
\end{equation}

In the case where the inputs contain both texts and images, a combiner module, notated as $g_{\tss{\theta}{c}}$, is typically needed to combine the features for tasks, such as composed image retrieval. The process can be defined as follows,
\begin{equation}
\tss{\mathbf{h}}{q} = 
\begin{cases}
    g_{\tss{\theta}{c}}(\tss{\mathbf{h}}{q-i}, \tss{\mathbf{h}}{q-t}), & \text{if } q = \{\tss{q}{i}, \tss{q}{t}\} \\
    \tss{\mathbf{h}}{q-i}, & \text{if } q = \tss{q}{i} \\
    \tss{\mathbf{h}}{q-t}, & \text{if } q = \tss{q}{t}
\end{cases}
\label{eq:comb}
\end{equation}
$\tss{\mathbf{h}}{q} \in \mathbb{R}^D $ is the embedding representing the query and the embedding of a candidate $\tss{\mathbf{h}}{c} \in \mathbb{R}^D $ can be obtained in the same way with identical $f$ and $g$. 

\paragraph{Optimization objective.}
Given a batch of pairs, $B = \{\{q_i, c_i\}\}^{|B|}_{i=1}$, where $q_i \in \mathcal{Q}$ and $c_i \in \mathcal{C}$ are termed query and the targeted candidate. As introduced in Equations~\eqref{eq:qc}, \eqref{eq:bienc} and \eqref{eq:comb}, we obtain embeddings as $\{\{\tss{\mathbf{h}}{q}_i, \tss{\mathbf{h}}{c}_i\}\}^{|B|}_{i=1}$. The set of parameters to be optimized is $\Theta = \{\tss{\theta}{t}, \tss{\theta}{i}, \tss{\theta}{c}\}$. Contrastive learning objective~\cite{oord2018representation} is used to optimize parameters $\Theta$ by minimizing the following InfoNCE loss,
\begin{equation}
    \mathcal{L} = -\frac{1}{|B|}\sum_{1 \leq i \leq |B|}\log{
    \frac{
        \exp(\tss{\mathbf{h}}{q}_i \cdot \tss{\mathbf{h}}{c}_i/\tau)
    }{
        \sum_{1 \leq j \leq |B|}\exp(\tss{\mathbf{h}}{q}_i \cdot \tss{\mathbf{h}}{c}_j/\tau)},
    }
\label{eq:loss}
\end{equation}
where $\tau$ is a temperature term controlling the discrimination to the negative samples.

\subsection{\modelfullname{}}
While conventional two-tower models have significantly transformed the landscape of image-to-text and text-to-image retrieval and perform well on simple text-image matching tasks, these models can still be limited because
\begin{itemize}[leftmargin=0.1in,itemsep=0in]
    \item The inability to jointly model visual and language data because of their separate encoding processes.
    \item The requirement for task-specific combiners to handle more complex tasks, such as composed image retrieval, due to the single-modal input and embeddings.
\end{itemize}

\paragraph{Unified encoding and embedding process.}
We propose \modelfullname{} (\modelname{}), which builds on an MLLM, \eg, ~\cite{beyer2024paligemma}, as its backbone encoder, unifying the encoding process for both queries and candidates. Given an input, either a query $q \in \mathcal{Q}$ or a candidate $c \in \mathcal{C}$, that may contain multi-modal information, we augment the token sequence by appending a special token \texttt{[Emb]} to its end. For example, for a query we form the augmented input $x_q = [q; \texttt{[Emb]}]$
which is processed by the MLLM’s shared Transformer encoder:
\begin{align}
\label{eq:emb_token_encoding}
\mathbf{e}_q = f_{\theta}(x_q) = \{\mathbf{e}_{q,1}, \mathbf{e}_{q,2}, \dots, \mathbf{e}_{q,N_q}\},
\end{align}
where $N_q$ is the total number of tokens in the augmented sequence. We then extract the hidden state corresponding to the \texttt{[Emb]} token as the final query embedding:
\begin{align}
\label{eq:emb_token_indexing}
\mathbf{h}_q \triangleq \mathbf{e}_{q,N_q} \in \mathbb{R}^D.
\end{align}
Similarly, for a candidate, we obtain its embedding by
$\mathbf{h}_c \triangleq \mathbf{e}_{c,N_c} \in \mathbb{R}^D$.
These embeddings $\mathbf{e}_q$ and $\mathbf{e}_c$ serve as the representations for the contrastive loss in Equation~\eqref{eq:loss}, which encourages corresponding query--candidate pairs to have similar embeddings while pushing apart those of unrelated pairs.

Although MLLMs are originally trained on generative tasks (e.g., visual question answering) that focus on next-token prediction, they are not naturally designed to extract discriminative representations for retrieval. To address this limitation, we first perform post-training adaptation on the backbone MLLM using large-scale paired data, followed by instruction tuning. Unlike the conventional two-tower model described in \S\ref{sec:pre}, this unified encoding process eliminates the need for an additional combiner—even when handling multi-modal queries such as in composed image retrieval.

\paragraph{Post-training Adaptation.}
In this stage, we fine-tune the backbone MLLM using the image-caption datasets to generate retrieval-specific embeddings. For each image-caption pair, we create two training instances by swapping roles: one instance treats the image as the query and the caption as the candidate, while the other reverses these roles. Each instance is processed by appending a special token (e.g., \texttt{[Emb]}) to the input sequence to designate the location for extracting the final embedding. The MLLM then encodes these augmented inputs, and a contrastive learning objective aligns embeddings from matching image–caption pairs while distinguishing non-matching pairs. This process ensures that both images and captions are effectively represented for retrieval tasks.

\paragraph{Instruction Tuning.}
Recent studies in text retrieval have explored integrating instructions into retrievers to better align with users' intentions~\cite{asai-etal-2023-task, su-etal-2023-one}. This challenge is more pronounced in Multi-modal retrieval, where instructive information can be presented in both textual and visual modalities for different tasks. This complexity necessitates the simultaneous processing of textual and visual instructive inputs during encoding. To further refine the MLLM for vision-language retrieval and better align it with human intent, we incorporate explicit task-specific instructions into the input sequences. In this stage, we leverage the instruction data from UniIR~\cite{wei2024uniir} to tune the unified MLLM for retrieval tasks. Specifically, for an input query $q$ (which may represent either visual or textual content) and its corresponding instruction $i$, we construct an augmented sequence by appending the instruction to the context along with a special token for embedding extraction:
\begin{align}
\label{eq:inst_format}
q' = [q; i; \texttt{[Emb]}].
\end{align}
Here, \texttt{[Emb]} marks the position from which the final embedding is extracted. The MLLM processes this combined sequence as a single input as in Equations~\eqref{eq:emb_token_encoding} and \eqref{eq:emb_token_indexing}, thereby jointly encoding the primary content and the instructive cues. The training objective (as defined in Equation~\eqref{eq:loss}) then aligns the embeddings of matching pairs while distinguishing those of non-matching pairs. This integrated learning approach enables the model to effectively interpret and execute retrieval tasks in accordance with human instructions.

\paragraph{Data sampling strategy.}
Our instruction data originates from multiple datasets and tasks, each exhibiting unbalanced data volumes. Since contrastive learning is sensitive to both intra-dataset and inter-dataset batch composition—relying on the effective mining of negative examples—we design a sampling scheme that carefully controls the number of datasets included in each batch. Empirically, we observe that limiting the number of datasets per batch improves overall performance. To achieve this, we sample the number of datasets per batch from a normal distribution (with rounding),
$N_d \sim \mathcal{N}(4, 1)$,
thereby ensuring that each batch contains a small, balanced subset of datasets. This strategy helps mitigate data imbalance while maintaining a rich set of negative samples, ultimately enhancing the robustness of the contrastive learning process.

\paragraph{Summary.}
Through the post-training adaptation and instruction tuning steps, we transform the MLLM into a powerful encoder for retrieval tasks. This adapted model excels in encoding multi-modal inputs into meaningful embeddings. Our approach leverages the MLLM's inherent ability to understand multi-modal instructive information, resulting in a unified encoding process that seamlessly handles various input types - be it text-only, image-only, or a combination of both.
\section{Experiments}
\label{sec:exp}
\subsection{Datasets}
\paragraph{CC3M~\cite{sharma2018conceptual}.} We performed the post-training adaption on the CC3M dataset \citep{sharma2018conceptual}, which consists of 3.3 million image-text pairs from the web. Using the $\operatorname{img2dataset}$ toolbox~\citep{beaumont-2021-img2dataset}, we download the dataset based on the provided URL-caption pairs, resulting in approximately 2.8 million image-text pairs due to some expired links.

\paragraph{M-BEIR~\cite{wei2024uniir}.} We instruction-tune the models on the M-BEIR dataset, which is a multimodal retrieval dataset encompassing eight tasks and ten datasets across domains like everyday imagery, fashion, Wikipedia, and news. It includes 1.5 million queries and 5.6 million retrieval candidates, despite being originally designed for various purposes. These include retrieval-focused datasets (e.g., OVEN~\cite{hu2023oven}, CIRR~\cite{liu2021cirr}, FashionIQ~\cite{wu2021fashioniq}), image-caption datasets (e.g., MS-COCO~\cite{lin2014microsoft}, Fashion200K~\cite{han2017fashion200k}), an image-similarity dataset (NIGHTS~
\cite{fu2023nights}), and retrieval-based VQA datasets (InfoSeek~\cite{chen2023infoseek}, WebQA~\cite{chang2022webqa}). For each dataset above, \citet{wei2024uniir} generated 4 instructions that describe a multimodal retrieval task by intent, domain, query modality, and target candidate modality. Thanks to its diverse input/output formats, MBEIR provides a suitable platform to train and evaluate multimodal retrieval systems.

\begin{table*}[ht!]
\small
\centering
\caption{Retrieval results on M-BEIR benchmark~\cite{wei2024uniir}.}
\vspace{-0.1in}
\label{tab:multitask_local}
\setlength{\tabcolsep}{1.5mm}{
\begin{tabular}{llcccccccca}
\toprule
 \multirow{2}{*}{\textbf{Task}} & \multirow{2}{*}{\textbf{Dataset}} &  \multicolumn{4}{c}{\textbf{SoTA Zero-Shot}} & \multicolumn{2}{c}{\textbf{Single-task FT}} & \multicolumn{3}{c}{\textbf{Multi-task (w/ instruction)}} \\
\cmidrule(lr){3-6} \cmidrule(lr){7-8} \cmidrule(lr){9-11}
& &CLIP & SigLIP & BLIP & BLIP2 &CLIP$_{\text{SF}}$ & BLIP$_{\text{FF}}$ & CLIP$_{\text{SF}}$ & BLIP$_{\text{FF}}$ & Ours\\
\midrule\midrule
$q_t \to c_t$ & WebQA & 36.2 & 39.8 & 44.9 & 38.6 & 81.7 & 67.5 & 84.1 & 79.2 & 88.7 \\ 
\midrule
$q_i \to c_i$ & NIGHTS & 26.1 & 28.9 & 27.4 & 25.4 & 33.5 & 30.4 & 31.1 & 31.7 & 27.8 \\ 
\midrule
\multicolumn{2}{l}{\textsc{Single-modal Average}} & 31.2	& 34.4 &	36.2 &	32 &	57.6 &	49.0 &	57.6 &	55.5 &	58.3 \\
\midrule\midrule
\multirow{3}{*}{$q_t \to c_i$} & VisualNews & 43.3 & 30.1 & 16.4 & 16.7 & 43.5 & 20.0 & 42.5 & 22.9 & 34.6 \\
 & MSCOCO & 61.1 & 75.7 & 74.4 & 63.8 & 80.4 & 77.3 & 80.7 & 79.5 & 78.5 \\
 & Fashion200K & 6.6 & 36.5 & 15.9 & 14.0 & 10.7 & 17.1 & 18.1 & 26.2 & 37.2 \\ 
\midrule
\multirow{3}{*}{$q_i \to c_t$} & VisualNews & 41.3 & 30.8 & 17.2 & 15.0 & 42.7 & 22.4 & 42.5 & 23.1 & 33.1 \\
 & MSCOCO & 79.0 & 88.2 & 83.2 & 80.0 & 89.8 & 86.0 & 91.8 & 90.8 & 90.0 \\
 & Fashion200K & 7.7 & 34.2 & 19.9 & 14.2 & 12.0 & 15.6 & 18.3 & 28.6 & 36.9 \\ 
\midrule
\multicolumn{2}{l}{\textsc{Cross-modal Average}} & 39.8 &	49.3 &	37.8 &	34.0 &	46.5 &	39.7 &	49.0 &	45.2	& 51.7 \\
\midrule\midrule
\multirow{2}{*}{$q_t \to$ ($c_i, c_t$)} & EDIS & 43.3 & 27.0 & 26.8 & 26.9 & 58.8 & 38.2 & 53.6 & 49.9 & 54.3 \\
 & WebQA & 45.1 & 43.5 & 20.3 & 24.5 & 76.3 & 67.8 & 78.3 & 78.1 & 82.4 \\ \midrule
\multirow{2}{*}{($q_i, q_t$) $\to c_t$} & OVEN & 24.2 & 29.7 & 16.1 & 12.2 & 45.4 & 33.8 & 46.0 & 42.7 & 46.0 \\
 & InfoSeek & 20.5 & 25.1 & 10.2 & 5.5 & 23.5 & 18.5 & 27.4 & 23.3 & 35.6 \\ \midrule
\multirow{2}{*}{($q_i, q_t$) $\to c_i$} & FashionIQ & 7.0 & 14.4 & 2.3 & 4.4 & 25.9 & 3.0 & 24.8 & 29.2 & 31.8 \\
 & CIRR & 13.2 & 22.7 & 10.6 & 11.8 & 52.0 & 13.9 & 44.6 & 50.7 & 54.0 \\
\midrule
\multicolumn{2}{l}{\textsc{Mixed-modal Average}} & 25.6	& 27.1 & 14.4 & 14.2 & 47.0 & 29.2 &	45.8 & 45.7 & 50.7 \\
\midrule\midrule
\multirow{2}{*}{($q_i, q_t$) $\to$ ($c_i, c_t$)} & OVEN & 38.8 & 41.7 & 27.4 & 27.3 & 66.2 & 49.9 & 68.7 & 56.5 & 72.7 \\
 & InfoSeek & 26.4 & 27.4 & 16.6 & 15.8 & 47.4 & 32.3 & 48.8 & 30.4 & 61.1 \\ 
\midrule
\multicolumn{2}{l}{\textsc{Multi-modal Average}} & 32.6 &	34.6 & 22.0 & 21.55 & 56.8 &	41.1 &	58.8 &	43.5 &	66.9 \\
\midrule\midrule
\multicolumn{2}{l}{\textsc{All Average}} & 32.5 & 37.2 & 26.8 & 24.8 & 49.4 & 37.1 & 50.1 & 46.4 & \textbf{54.0} \\
\bottomrule
\end{tabular}
}
\end{table*}

\subsection{Training setups}
For most of the experiments in this paper, we default to PaliGemma~\cite{beyer2024paligemma} as the choice of MLLMs because of its relatively compact size ($\sim$3B) and competitive performance on various vision-language understanding benchmarks. Following the original recipe, the image input of the model is simply scaled to size $256\times 256$ and fed into a SigLIP vision encoder (a ViT with patch size of $16\times 16$) to obtain 256 vision tokens; the text input is tokenized by the sentence piece tokenizer. For training efficiency and reducing GPU memory consumption, we truncate the input token when the total number of tokens is larger than 384, which means the maximal length of textual tokens (including those of textual instruction) is 128 when the input contains an image or 378 otherwise.

For the post-training adaption, we train the MLLMs on CC3M~\cite{sharma2018conceptual} datasets for 1 epoch using Low-Rank Adapters (LoRA)~\cite{hu2022lora} with $r = 128, \alpha = 256$, and a dropout probability 0.05. We use the AdamW~\cite{loshchilov2017decoupled} optimizer with a learning rate of 2e-4, a batch size of 2048, and no weight decay for the LoRA training. The learning rate is linearly warmed-up for the first 3\% of training to the specified value and then decayed using a cosine annealing schedule~\cite{loshchilov2016sgdr}.

For the instruction-tuning stage, we first merge the LoRA of the first stage to the base MLLM and then reinitialize and train a new of new LoRA based on the merged weights. We use  $r = 256, \alpha = 512$, and a dropout probability 0.3 for LoRA at this stage because we find it beneficial to use more parameters to enhance the instruction-following capability (see Tab.~\ref{tab:lora_params}). We train the LoRA with a batch size of 1024, no weight decay, and a learning rate of 2e-4 which is warmed-up and decayed as in the first stage.

\begin{table*}[ht!]
\small
\centering
\caption{Conditional Retrieval on GeneCIS benchmark~\cite{Vaze2023GeneCIS}.}
\label{tab:cond_retr}
\vspace{-0.1in}
\resizebox{\textwidth}{!}{%
\setlength{\tabcolsep}{1mm}{
\begin{tabular}{lccccccccccccccc}
\toprule
\multicolumn{1}{l}{\multirow{2}{*}{\textbf{Method}}} & \multicolumn{3}{c}{\textbf{Focus Attribute}} & \multicolumn{3}{c}{\textbf{Change Attribute}} & \multicolumn{3}{c}{\textbf{Focus Object}} & \multicolumn{3}{c}{\textbf{Change Object}} & \textbf{Avg} \\ 
\cmidrule(lr){2-4} \cmidrule(lr){5-7} \cmidrule(lr){8-10} \cmidrule(lr){11-13} \cmidrule(lr){14-14}
\multicolumn{1}{c}{} & R@1 & R@2 & R@3 & R@1 & R@2 & R@3 & R@1 & R@2 & R@3 & R@1 & R@2 & R@3 & R@1 \\ 
\midrule
Pic2Word \cite{Saito2023Pic2Word} & 12.5 & 23.4 & 33.7 & 11.7 & 21.9 & 30.9 & 9.9 & 19.3 & 27.4 & 8.6 & 18.2 & 26.1 & 10.7 \\
SEARLE \cite{Baldrati2023CIRCO} & 16.3 & 29.4 & 40.7 & 16.2 & 27.3 & 35.5 & 10.8 & 18.2 & 27.9 & 8.3 & 15.6 & 25.8 & 12.9 \\
CompoDiff \cite{Gu2023CompoDiff} & 14.3 & 26.7 & 38.4 & 19.7 & 28.8 & 37.4 & 9.2 & 19.1 & 25.8 & 18.7 & 31.7 & 40.2 & 15.5 \\
CIReVL \cite{Karthik2023CIReVL} & 20.5 & 34.0 & 44.5 & 16.1 & 28.6 & 39.4 & 14.7 & 25.2 & 33.0 & 18.1 & 31.2 & 41.0 & 17.4 \\
LinCIR \cite{Gu2023LinCIR} & 19.1 & 33.0 & 42.3 & 17.6 & 30.2 & 38.1 & 10.1 & 19.1 & 28.1 & 7.9 & 16.3 & 25.7 & 13.7 \\ 
MagicLens~\cite{zhang2024magiclens} & 16.6 & 28.7 & 39.3 & 16.0 & 27.5 & 36.5 & 15.7 & 27.6 & 37.3 & 18.7 & 31.7 & 40.2 & 16.7 \\ 
\midrule
CLIP$_{\text{SF}}$ \cite{wei2024uniir} & 21.1 & 33.9 & 44.6 & 15.1 & 27.6 & 37.8 & 15.0 & 25.3 & 35.0 & 13.6 & 24.8 & 35.7 & 16.2 \\
BLIP$_{\text{FF}}$ \cite{wei2024uniir} & 19.4&32.3&44.0&15.8&26.9&36.0&18.0&28.4&37.0&18.5&29.4&39.1 & 17.9\\
\midrule
\rowcolor{Gray} Ours & 18.9 &	29.6 &	40.7 &	15.7 &	28.0 &	36.7&	21.5 &	32.7 &	40.5 &	24.1 &	37.9 & 48.4 & \textbf{20.1} \\
\bottomrule
\end{tabular}
}
}
\end{table*}

\subsection{Evaluation setups}
We mainly evaluate \modelname{} on the M-BEIR benchmark~\cite{wei2024uniir} because it contains a diverse set of input and target modalities and provides large-scale query and candidate sets (190K queries and 5.6M candidates) for reliable evaluations. We adopt the settings that perform retrieval from a task-specific pool provided by the original dataset, enabling comparison with non-instruction-tuned retrievers. We report the Recall@5 for all the datasets except FashionIQ and Fashion 200K, where Recall@10 is used following \citet{wu2021fashioniq}.

In addition, we also evaluate \modelname{} on conditional image similarity, which measures the capability of models not only in encoding the content of the query but also in understanding users' intent or conditions. We use the GeneCIS benchmark~\cite{Vaze2023GeneCIS}, an image-to-image retrieval task conditioned on several keywords. GeneCIS consists of four sub-tasks about focusing or changing on a specific attribute or object. For instance, for the sub-task about focusing on an object, the models need to find the most relevant image with the same object (specified in the condition) as the query.

\subsection{Single-modal and cross-modal retrieval}
We begin by evaluating our method in single-modal and cross-modal settings, where both the query and candidate consist of a single modality. Although the baseline models are specifically designed to handle single-modal inputs, our multi-modal input method slightly outperforms them in single-modal retrieval, achieving an average score of 58.3 compared to 57.6 for CLIP$_{\text{SF}}$ and 49.0 for BLIP$_{\text{FF}}$. Similarly, in cross-modal retrieval, our method attains an average score of 51.7, surpassing single-task fine-tuned models (46.5 for CLIP$_{\text{SF}}$ and 39.7 for BLIP$_{\text{FF}}$) and multi-task models (49.0). These results indicate that even in tasks where two-tower-based methods typically excel, our unified multi-modal approach delivers competitive and even slightly superior performance.

\subsection{Mixed-modal and multi-modal retrieval}
We further evaluate our model in mixed- and multi-modal settings, where both queries and candidates can be arbitrary combinations of images and text. For mixed-modal retrieval, \modelname{} achieves an average score of 50.7, significantly outperforming multi-task baselines (45.8 and 45.7) and single-task BLIP$_{\text{FF}}$ (29.2). In multi-modal tasks, our model achieves an average score of 66.9, surpassing both single-task FT (56.8) and multi-task models (58.8) by 8 points. These results demonstrate that, unlike two-tower-based baselines which struggle to interpret multi-modal inputs, our approach effectively integrates cross-modal and multi-task signals, obtaining superior performance in complex multi-modal retrieval scenarios. These experiments reiterate the importance of unified models, like \modelname{}, for vision-language retrieval.

\begin{table}[t]
    \centering
    \small
    \caption{Influence of two-stage training.}
    \label{tab:2stage_training}
    \vspace{-0.1in}
    \setlength{\tabcolsep}{1.5mm}{
    \begin{tabular}{cccccc}
    \toprule
    \multirow{2}{*}{\textbf{Stage 1}} & \multirow{2}{*}{\textbf{Stage 2}} & \multicolumn{4}{c}{\textbf{Retrieval recall}} \\
    \cmidrule(lr){3-6}
    & & Single & Cross & Mixed & Average \\
    \midrule
    \xmark & \xmark & 6.7 & 0.1 & 0.0 & 1.4 \\
    \cmark & \xmark & 59.3 & 54.3 & 6.7 & 36.3 \\
    \xmark & \cmark & 84.8 & 80.5 & 44.3 & 66.9 \\
    \rowcolor{color_ours} \cmark & \cmark & 88.2 & 84.3 & 42.3 & 68.3 \\
    \bottomrule
    \end{tabular}
    }
\end{table}

\subsection{Conditional retrieval}
Following \citet{Vaze2023GeneCIS}, we report the Recall@K, K = $\{1, 2, 3\}$ for all four sub-tasks, as well as the averaged Recall@1 in Tab.~\ref{tab:cond_retr}. We conduct in-depth comparisons with various state-of-the-art methods, all following the two-tower fashion and potentially with a combiner. This includes 1) Pic2Word~\cite{Saito2023Pic2Word}, SEARLE~\cite{Baldrati2023CIRCO} and LinCIR~\cite{Gu2023LinCIR} that map images into a special text token inserted to the condition prompts; 2) CIReVL~\cite{Karthik2023CIReVL} that first captions the image and then merges the caption and the condition using LLMs to a textual query; 3) CompoDiff~\cite{Gu2023CompoDiff} and MagicLens~\cite{zhang2024magiclens} which curate or synthesize composed image retrieval data for training a two-tower model; and 4) UniIR variants (CLIP$_{\textbf{SF}}$ and BLIP$_{\text{FF}}$) that are trained on the same data as ours.
From the table, we observe that \modelname{} delivers competitive performance across all tasks compared to SOTA methods, with exceptional results in object-centric conditions. \modelname{} achieves an averaged Recall@1 of 20.1, significantly outperforming all other methods by a large margin, despite not relying on any task-specific design for conditional retrieval. This underscores the effectiveness of using a unified model to jointly comprehend vision and language information. Additionally, although BLIP$_{\text{FF}}$ shows relatively competitive performance with an average Recall@1 of 17.9, it still falls significantly short of the robustness demonstrated by our approach, especially in subtasks involving object modification or composition. This reinforces that the improvements achieved by \modelname{} are primarily due to its architectural design rather than any advantage from the data.

\begin{table}[t]
    \centering
    \small
    \caption{Influence of data sampling strategy.}
    \label{tab:data_sampling}
    \vspace{-0.1in}
    \begin{tabular}{ccccc}
    \toprule
     \multirow{2}{*}{\textbf{\#Dataset/Batch}} & \multicolumn{4}{c}{\textbf{Retrieval recall}} \\
    \cmidrule(lr){2-5}
    & Single & Cross & Mixed & Average \\
    \midrule
    N/A & 81.1 & 78.4 & 37.9 & 62.8 \\
    2 & 75.4 & 81.5 & 39.5 & 63.4 \\
    4 & 84.5 & 82.0 & 41.3 & 66.2 \\
    8 & 81.8 & 77.9 & 40.9 & 63.9 \\
    \rowcolor{color_ours} $\mathcal{N}(4, 1)$ & 84.8 & 80.5 & 44.3 & 66.9 \\
    \bottomrule
    \end{tabular}
\end{table}

\begin{table}[t]
    \centering
    \small
    \caption{Influence of the hyper-parameters rank ($r$), $\alpha$, and dropout probability $d$ in LoRA.}
    \label{tab:lora_params}
    \vspace{-0.1in}
    \setlength{\tabcolsep}{1.5mm}{
    \begin{tabular}{ccccccc}
    \toprule
     \multicolumn{3}{c}{\textbf{LoRA hyper-param.}} & \multicolumn{4}{c}{\textbf{Retrieval recall}} \\
     \cmidrule(lr){1-3}
     \cmidrule(lr){4-7}
     $r$ & $\alpha$ & $d$ 
     & Single & Cross & Mixed & Average \\
    \midrule
    16 & 32 & 0.05 & 78.3 & 77.7 & 35.9 & 61.1 \\
    128 & 256 & 0.05 & 81.5 & 80.2 & 38.3 & 63.7 \\
    128 & 256 &  0.3 & 81.9 & 83.8 & 37.3 & 64.8 \\
    \rowcolor{color_ours} 256 & 512 & 0.3 & 84.8 & 80.5 & 44.3 & 66.9 \\
    \bottomrule
    \end{tabular}
    }
\end{table}

\subsection{Ablation analysis}

\paragraph{Impact of Two-stage Training.} In Tab.~\ref{tab:2stage_training}, we study the impact of the two-stage training on retrieval performance using a subset of M-BEIR.
Without any training, the performance of the original MLLMs is no better than random guessing, indicating the necessity of carefully designed adaption steps. Applying stage 1 alone yields a significant performance improvement, but falls behind stage 2 training alone, which is reasonable considering the discrepancy between the CC3M and M-BEIR. The combination of both stages results in the best retrieval recall especially in the case of single-modal retrieval, suggesting the benefits of the post-training adaption for tasks involving only single-modality query/candidate.

\paragraph{Benefits of data sampling.} 
Tab.~\ref{tab:data_sampling} investigates how different data sampling strategies influence retrieval performance. Training without any sampling strategy underperforms (62.8 average recall) those with sampling operations, emphasizing the necessity of batch diversity. As the dataset count per batch increases, performance improves up to a peak at 4 datasets per batch. The use of Gaussian sampling $\mathcal{N}(4, 1)$ further improves this result, achieving an averaged score of 66.9. This suggests that properly balancing the data source within batches benefits generalization.

\paragraph{Impact of the LoRA hyper-parameters and batch size.} 
Tabs.~\ref{tab:lora_params} and \ref{tab:training_bs} explore the importance of LoRA hyper-parameters and batch size on retrieval performance. For LoRA, larger rank and scaling factors ($r = 256$, $\alpha = 512$) consistently yield performance gains, achieving the best retrieval recall of 66.9. Additionally, moderate dropout regularization (0.3) outperforms smaller rates (0.05), suggesting that balancing parameter complexity and overfitting is crucial for robust performance. Meanwhile, batch size plays a critical role, with a larger batch size of 1024 reaching 66.9 averaged score, showing that batch size is particularly impactful as larger batches typically lead to better negative sampling and stronger representation learning. However, due to GPU memory limitations, batch sizes beyond 1024 could not be tested. We would expect performance to further improve with larger batch sizes.

\paragraph{Scaling the number of training epochs.} 
Tab.~\ref{tab:training_ep} analyzes how the number of training epochs affects retrieval performance. From the table, we can see that the performance considerably increases from 67.6 at 2 epochs to 68.6 at 3 epochs. Beyond this, the performance plateaus, as both 3 and 5 epochs yield the same overall score (68.6 recall). These results suggest that training saturation occurs beyond 3 epochs, where additional epochs provide diminishing returns on retrieval effectiveness. We default to 3 epochs in our experiments for a good trade-off between performance and training efficiency.

\begin{table}[t]
    \centering
    \small
    \caption{Influence of the number of training batch size.}
    \label{tab:training_bs}
    \vspace{-0.1in}
    \begin{tabular}{ccccc}
    \toprule
     \multirow{2}{*}{\textbf{Batch Size}} & \multicolumn{4}{c}{\textbf{Retrieval recall}} \\
     \cmidrule(lr){2-5}
     & Single & Cross & Mixed & Average \\
    \midrule
    256 & 80.2 & 77.8 & 41.4 & 63.7 \\
    512 & 77.5 & 82.9 & 41.1 & 65.1 \\
    \rowcolor{color_ours} 1024 & 84.8 & 80.5 & 44.3 & 66.9 \\
    \bottomrule
    \end{tabular}
\end{table}

\begin{table}[t]
    \centering
    \small
    \caption{Influence of the number of training epochs.}
    \label{tab:training_ep}
    \vspace{-0.1in}
    \begin{tabular}{ccccc}
    \toprule
    \multirow{2}{*}{\textbf{Num. Epochs}} & \multicolumn{4}{c}{\textbf{Retrieval recall}} \\
     \cmidrule(lr){2-5}
     & Single & Cross & Mixed & Average \\
    \midrule
    2 & 87.1 & 80.8 & 44.6 & 67.6 \\
    \rowcolor{color_ours} 3 & 88.7 & 84.3 & 42.9 & 68.6 \\
    5 & 86.9 & 84.2 & 44.0 & 68.6 \\
    \bottomrule
    \end{tabular}
\end{table}

\begin{table*}[t]
\small
\centering
\caption{Retrieval results on M-BEIR$_{\mathrm{global}}$}
\label{tab:multitask}
\resizebox{\textwidth}{!}{%
\begin{tabular}{llccccca}
\toprule
 &  & \textbf{Zero-shot} & \multicolumn{2}{c}{\textbf{Multi-task (w/ instruction)}} & \multicolumn{3}{c}{\textbf{Multi-task (w/o instruction)}}  \\ \cmidrule(lr){3-3} \cmidrule(lr){4-5} \cmidrule(lr){6-8}
\textbf{Task} & \textbf{Dataset} & BLIP2 & CLIP$_{\text{SF}}$ & BLIP$_{\text{FF}}$ & CLIP$_{\text{SF}}$ & BLIP$_{\text{FF}}$ & Ours \\ 
\midrule
\multirow{3}{*}{1. $q_t \to c_i$} & VisualNews & 0.0 & 12.7 & 8.3 & 42.2 & 22.5 & 31.5 \\
 & MSCOCO & 0.0 & 27.3 & 27.7 & 71.4 & 65.3 & 62.1 \\
 & Fashion200K & 0.0 & 5.9 & 9.0 & 18.0 & 26.1 & 35.6 \\ \midrule
2. $q_t \to c_t$ & WebQA & 35.2 & 82.3 & 76.1 & 83.5 & 78.5 & 87.6 \\ \midrule
\multirow{2}{*}{3. $q_t \to$ ($c_i, c_t$)} & EDIS & 0.0 & 41.1 & 36.0 & 52.7 & 49.3 & 51.7 \\
 & WebQA & 0.0 & 68.2 & 74.7 & 77.5 & 77.1 & 81.0 \\ \midrule
\multirow{3}{*}{4. $q_i \to c_t$} & VisualNews & 0.0 & 12.1 & 4.9 & 38.8 & 21.1 & 30.3 \\
 & MSCOCO & 0.0 & 84.6 & 76.9 & 91.4 & 89.8 & 89.0 \\
 & Fashion200K & 0.0 & 1.2 & 3.6 & 18.2 & 27.4 & 30.9 \\ \midrule
5. $q_i \to c_i$ & NIGHTS & 24.0 & 31.0 & 31.3 & 39.5 & 31.6 & 27.8 \\ \midrule
\multirow{2}{*}{6. ($q_i, q_t$) $\to c_t$} & OVEN & 0.0 & 36.8 & 37.7 & 22.2 & 39.5 & 42.4 \\
 & InfoSeek & 0.0 & 18.3 & 17.8 & 24.6 & 19.8 & 31.9 \\ \midrule
\multirow{2}{*}{7. ($q_i, q_t$) $\to c_i$} & FashionIQ & 3.9 & 22.8 & 28.1 & 43.1 & 28.9 & 31.1 \\
 & CIRR & 6.2 & 32.0 & 45.1 & 59.8 & 48.3 & 50.4 \\ \midrule
\multirow{2}{*}{8. ($q_i, q_t$) $\to$ ($c_i, c_t$)} & OVEN & 13.8 & 58.7 & 51.6 & 44.3 & 55.9 & 69.1 \\
 & InfoSeek & 11.4 & 42.3 & 25.4 & 44.3 & 26.2 & 57.4 \\ \midrule
 & Average & 5.9 & 36.1 & 34.6 & 47.4 & 44.2 & 50.6 \\
\bottomrule
\end{tabular}
}
\end{table*}

\section{Conclusion}
In this work, we introduced \modelname{}, a one-tower multi-modal retrieval framework that integrates fusion directly into the encoding process. By adapting MLLMs into effective encoders through post-training adaptation and instruction-tuning, \modelname{} captures fine-grained interactions between visual and textual cues from the ground up. Our extensive evaluations demonstrate that \modelname{} achieves moderate gains in standard single-modal and cross-modal retrieval, its performance improvements become particularly pronounced in complex scenarios involving multi-modal or conditional queries, where the modality fusion is required.
Overall, the findings underscore the superiority of joint fusion and encoding for advanced multi-modal retrieval applications with inputs requiring complex multi-modal understanding.

\section*{Limitations}
This work is built upon large-scale pre-trained models rather than developing two-tower and one-tower architectures from scratch. Although these pre-trained models have been widely adopted in the community, their use introduces two primary limitations. First, the influence of the pre-trained models cannot be fully isolated—since both the pre-training approach and the underlying data have not been entirely publicly disclosed, their contributions remain a confounding factor. Second, there are efficiency concerns, particularly for retrieval tasks that demand fast online inference. These issues could be mitigated by further advances in efficient large models and the development of more streamlined backbone architectures.

\section*{Ethical Statement}
This research focuses on the daily task of information retrieval, which in itself does not pose ethical concerns. Our approach employs an encoding model to compress information, thereby mitigating the risk of inappropriate data generation. All datasets and pre-trained checkpoints used in this study are publicly available with free use for research and remain unaltered. However, as is common in much of today’s AI research, the performance of large AI models is not yet fully understood. Our evaluation is limited to academic benchmarks, and we do not endorse their deployment in practical applications at this stage.

\appendix
\section*{Appendix}
\label{sec:appendix}

\section{Additional Experimental Results}
We also adopt an alternative retrieval setting as described in \cite{wei2024uniir}, which conducts retrieval from a pool of 5.6 million candidates aggregated from eight tasks across ten M-BEIR datasets. As demonstrated in Table~\ref{tab:multitask}, despite the significantly varied evaluation settings, \modelname{} not only achieves moderate improvements over baselines in single-modal and cross-modal retrieval but also delivers substantial gains in mixed-modal and multi-modal scenarios—settings that are increasingly relevant in our multi-modal content-rich world. These results further corroborate our findings in \S\ref{sec:exp} and underscore the advantages of the Joint Fusion and Encoding paradigm.

\bibliography{anthology,custom}

\begin{thebibliography}{40}
\providecommand{\natexlab}[1]{#1}

\bibitem[{Asai et~al.(2023)Asai, Schick, Lewis, Chen, Izacard, Riedel, Hajishirzi, and Yih}]{asai-etal-2023-task}
Akari Asai, Timo Schick, Patrick Lewis, Xilun Chen, Gautier Izacard, Sebastian Riedel, Hannaneh Hajishirzi, and Wen-tau Yih. 2023.
\newblock \href {https://doi.org/10.18653/v1/2023.findings-acl.225} {Task-aware retrieval with instructions}.
\newblock In \emph{Findings of the Association for Computational Linguistics: ACL 2023}, pages 3650--3675, Toronto, Canada. Association for Computational Linguistics.

\bibitem[{Baldrati et~al.(2023)Baldrati, Agnolucci, Bertini, and Del~Bimbo}]{Baldrati2023CIRCO}
Alberto Baldrati, Lorenzo Agnolucci, Marco Bertini, and Alberto Del~Bimbo. 2023.
\newblock Zero-shot composed image retrieval with textual inversion.
\newblock In \emph{Proceedings of the International Conference on Computer Vision}.

\bibitem[{Baldrati et~al.(2022)Baldrati, Bertini, Uricchio, and Del~Bimbo}]{9879245}
Alberto Baldrati, Marco Bertini, Tiberio Uricchio, and Alberto Del~Bimbo. 2022.
\newblock \href {https://doi.org/10.1109/CVPR52688.2022.02080} {Effective conditioned and composed image retrieval combining clip-based features}.
\newblock In \emph{2022 IEEE/CVF Conference on Computer Vision and Pattern Recognition (CVPR)}, pages 21434--21442.

\bibitem[{Beaumont(2021)}]{beaumont-2021-img2dataset}
Romain Beaumont. 2021.
\newblock img2dataset: Easily turn large sets of image urls to an image dataset.
\newblock \url{https://github.com/rom1504/img2dataset}.

\bibitem[{Beyer et~al.(2024)Beyer, Steiner, Pinto, Kolesnikov, Wang, Salz, Neumann, Alabdulmohsin, Tschannen, Bugliarello et~al.}]{beyer2024paligemma}
Lucas Beyer, Andreas Steiner, Andr{\'e}~Susano Pinto, Alexander Kolesnikov, Xiao Wang, Daniel Salz, Maxim Neumann, Ibrahim Alabdulmohsin, Michael Tschannen, Emanuele Bugliarello, et~al. 2024.
\newblock Paligemma: A versatile 3b vlm for transfer.
\newblock \emph{arXiv preprint arXiv:2407.07726}.

\bibitem[{Chang et~al.(2022)Chang, Narang, Suzuki, Cao, Gao, and Bisk}]{chang2022webqa}
Yingshan Chang, Mridu Narang, Hisami Suzuki, Guihong Cao, Jianfeng Gao, and Yonatan Bisk. 2022.
\newblock Webqa: Multihop and multimodal qa.
\newblock In \emph{Proceedings of the IEEE/CVF conference on computer vision and pattern recognition}, pages 16495--16504.

\bibitem[{Chen et~al.(2023)Chen, Hu, Luan, Sun, Changpinyo, Ritter, and Chang}]{chen2023infoseek}
Yang Chen, Hexiang Hu, Yi~Luan, Haitian Sun, Soravit Changpinyo, Alan Ritter, and Ming-Wei Chang. 2023.
\newblock Can pre-trained vision and language models answer visual information-seeking questions?
\newblock In \emph{Proceedings of the 2023 Conference on Empirical Methods in Natural Language Processing}, pages 14948--14968.

\bibitem[{Fu et~al.(2023)Fu, Tamir, Sundaram, Chai, Zhang, Dekel, and Isola}]{fu2023nights}
Stephanie Fu, Netanel~Y Tamir, Shobhita Sundaram, Lucy Chai, Richard Zhang, Tali Dekel, and Phillip Isola. 2023.
\newblock Dreamsim: learning new dimensions of human visual similarity using synthetic data.
\newblock In \emph{Advance on Neural Information Processing Systems}, pages 50742--50768.

\bibitem[{Gao et~al.(2021)Gao, Yao, and Chen}]{gao-etal-2021-simcse}
Tianyu Gao, Xingcheng Yao, and Danqi Chen. 2021.
\newblock \href {https://doi.org/10.18653/v1/2021.emnlp-main.552} {{S}im{CSE}: Simple contrastive learning of sentence embeddings}.
\newblock In \emph{Proceedings of the 2021 Conference on Empirical Methods in Natural Language Processing}, pages 6894--6910, Online and Punta Cana, Dominican Republic. Association for Computational Linguistics.

\bibitem[{Gao et~al.(2023)Gao, Xiong, Gao, Jia, Pan, Bi, Dai, Sun, and Wang}]{gao2023retrieval}
Yunfan Gao, Yun Xiong, Xinyu Gao, Kangxiang Jia, Jinliu Pan, Yuxi Bi, Yi~Dai, Jiawei Sun, and Haofen Wang. 2023.
\newblock Retrieval-augmented generation for large language models: A survey.
\newblock \emph{arXiv preprint arXiv:2312.10997}.

\bibitem[{Gu et~al.(2024)Gu, Chun, Kim, , Kang, and Yun}]{Gu2023LinCIR}
Geonmo Gu, Sanghyuk Chun, Wonjae Kim, , Yoohoon Kang, and Sangdoo Yun. 2024.
\newblock Language-only training of zero-shot composed image retrieval.
\newblock In \emph{Proceedings of the IEEE Conference on Computer Vision and Pattern Recognition}.

\bibitem[{Gu et~al.(2023)Gu, Chun, Kim, Jun, Kang, and Yun}]{Gu2023CompoDiff}
Geonmo Gu, Sanghyuk Chun, Wonjae Kim, HeeJae Jun, Yoohoon Kang, and Sangdoo Yun. 2023.
\newblock Compodiff: Versatile composed image retrieval with latent diffusion.
\newblock \emph{arXiv preprint arXiv:2303.11916}.

\bibitem[{Han et~al.(2017)Han, Wu, Huang, Zhang, Zhu, Li, Zhao, and Davis}]{han2017fashion200k}
Xintong Han, Zuxuan Wu, Phoenix~X Huang, Xiao Zhang, Menglong Zhu, Yuan Li, Yang Zhao, and Larry~S Davis. 2017.
\newblock Automatic spatially-aware fashion concept discovery.
\newblock In \emph{Proceedings of the IEEE International Conference on Computer Vision}, pages 1463--1471.

\bibitem[{Hu et~al.(2022)Hu, Wallis, Allen-Zhu, Li, Wang, Wang, Chen et~al.}]{hu2022lora}
Edward~J Hu, Phillip Wallis, Zeyuan Allen-Zhu, Yuanzhi Li, Shean Wang, Lu~Wang, Weizhu Chen, et~al. 2022.
\newblock Lora: Low-rank adaptation of large language models.
\newblock In \emph{International Conference on Learning Representations}.

\bibitem[{Hu et~al.(2023)Hu, Luan, Chen, Khandelwal, Joshi, Lee, Toutanova, and Chang}]{hu2023oven}
Hexiang Hu, Yi~Luan, Yang Chen, Urvashi Khandelwal, Mandar Joshi, Kenton Lee, Kristina Toutanova, and Ming-Wei Chang. 2023.
\newblock Open-domain visual entity recognition: Towards recognizing millions of wikipedia entities.
\newblock In \emph{Proceedings of the IEEE/CVF International Conference on Computer Vision}, pages 12065--12075.

\bibitem[{Jang et~al.(2023)Jang, Kong, Jeon, Kim, and Kwak}]{jang2023unifying}
Jiho Jang, Chaerin Kong, Donghyeon Jeon, Seonhoon Kim, and Nojun Kwak. 2023.
\newblock Unifying vision-language representation space with single-tower transformer.
\newblock In \emph{Proceedings of the AAAI Conference on Artificial Intelligence}, volume~37, pages 980--988.

\bibitem[{Jiang et~al.(2024)Jiang, Meng, Yang, Yavuz, Zhou, and Chen}]{jiang2024vlm2vec}
Ziyan Jiang, Rui Meng, Xinyi Yang, Semih Yavuz, Yingbo Zhou, and Wenhu Chen. 2024.
\newblock Vlm2vec: Training vision-language models for massive multimodal embedding tasks.
\newblock \emph{arXiv preprint arXiv:2410.05160}.

\bibitem[{Karthik et~al.(2024)Karthik, Roth, Mancini, and Akata}]{Karthik2023CIReVL}
Shyamgopal Karthik, Karsten Roth, Massimiliano Mancini, and Zeynep Akata. 2024.
\newblock Vision-by-language for training-free compositional image retrieval.
\newblock In \emph{International Conference on Learning Representation}.

\bibitem[{Koukounas et~al.(2024)Koukounas, Mastrapas, Wang, Akram, Eslami, G{\"u}nther, Mohr, Sturua, Martens, Wang et~al.}]{koukounas2024jina}
Andreas Koukounas, Georgios Mastrapas, Bo~Wang, Mohammad~Kalim Akram, Sedigheh Eslami, Michael G{\"u}nther, Isabelle Mohr, Saba Sturua, Scott Martens, Nan Wang, et~al. 2024.
\newblock jina-clip-v2: Multilingual multimodal embeddings for text and images.
\newblock \emph{arXiv preprint arXiv:2412.08802}.

\bibitem[{Lewis et~al.(2020)Lewis, Perez, Piktus, Petroni, Karpukhin, Goyal, K{\"u}ttler, Lewis, Yih, Rockt{\"a}schel et~al.}]{lewis2020retrieval}
Patrick Lewis, Ethan Perez, Aleksandra Piktus, Fabio Petroni, Vladimir Karpukhin, Naman Goyal, Heinrich K{\"u}ttler, Mike Lewis, Wen-tau Yih, Tim Rockt{\"a}schel, et~al. 2020.
\newblock Retrieval-augmented generation for knowledge-intensive nlp tasks.
\newblock \emph{Advances in Neural Information Processing Systems}, 33:9459--9474.

\bibitem[{Li and Tang(2024)}]{li2024multimodal}
Songtao Li and Hao Tang. 2024.
\newblock Multimodal alignment and fusion: A survey.
\newblock \emph{arXiv preprint arXiv:2411.17040}.

\bibitem[{Lin et~al.(2024)Lin, Lee, Shoeybi, Lin, Catanzaro, and Ping}]{lin2024mm}
Sheng-Chieh Lin, Chankyu Lee, Mohammad Shoeybi, Jimmy Lin, Bryan Catanzaro, and Wei Ping. 2024.
\newblock Mm-embed: Universal multimodal retrieval with multimodal llms.
\newblock \emph{arXiv preprint arXiv:2411.02571}.

\bibitem[{Lin et~al.(2014)Lin, Maire, Belongie, Hays, Perona, Ramanan, Doll{\'a}r, and Zitnick}]{lin2014microsoft}
Tsung-Yi Lin, Michael Maire, Serge Belongie, James Hays, Pietro Perona, Deva Ramanan, Piotr Doll{\'a}r, and C~Lawrence Zitnick. 2014.
\newblock Microsoft coco: Common objects in context.
\newblock In \emph{Computer Vision--ECCV 2014: 13th European Conference, Zurich, Switzerland, September 6-12, 2014, Proceedings, Part V 13}, pages 740--755. Springer.

\bibitem[{Liu et~al.(2023)Liu, Xiong, Lv, Liu, and Yu}]{liu2023universal}
Zhenghao Liu, Chenyan Xiong, Yuanhuiyi Lv, Zhiyuan Liu, and Ge~Yu. 2023.
\newblock \href {https://openreview.net/forum?id=PQOlkgsBsik} {Universal vision-language dense retrieval: Learning a unified representation space for multi-modal retrieval}.
\newblock In \emph{The Eleventh International Conference on Learning Representations}.

\bibitem[{Liu et~al.(2021{\natexlab{a}})Liu, Rodriguez-Opazo, Teney, and Gould}]{Liu_2021_ICCV}
Zheyuan Liu, Cristian Rodriguez-Opazo, Damien Teney, and Stephen Gould. 2021{\natexlab{a}}.
\newblock Image retrieval on real-life images with pre-trained vision-and-language models.
\newblock In \emph{Proceedings of the IEEE/CVF International Conference on Computer Vision (ICCV)}, pages 2125--2134.

\bibitem[{Liu et~al.(2021{\natexlab{b}})Liu, Rodriguez-Opazo, Teney, and Gould}]{liu2021cirr}
Zheyuan Liu, Cristian Rodriguez-Opazo, Damien Teney, and Stephen Gould. 2021{\natexlab{b}}.
\newblock Image retrieval on real-life images with pre-trained vision-and-language models.
\newblock In \emph{Proceedings of the IEEE/CVF International Conference on Computer Vision}, pages 2125--2134.

\bibitem[{Loshchilov and Hutter(2016)}]{loshchilov2016sgdr}
Ilya Loshchilov and Frank Hutter. 2016.
\newblock Sgdr: Stochastic gradient descent with warm restarts.
\newblock \emph{arXiv preprint arXiv:1608.03983}.

\bibitem[{Loshchilov and Hutter(2017)}]{loshchilov2017decoupled}
Ilya Loshchilov and Frank Hutter. 2017.
\newblock Decoupled weight decay regularization.
\newblock \emph{arXiv preprint arXiv:1711.05101}.

\bibitem[{Oord et~al.(2018)Oord, Li, and Vinyals}]{oord2018representation}
Aaron van~den Oord, Yazhe Li, and Oriol Vinyals. 2018.
\newblock Representation learning with contrastive predictive coding.
\newblock \emph{arXiv preprint arXiv:1807.03748}.

\bibitem[{Radford et~al.(2021)Radford, Kim, Hallacy, Ramesh, Goh, Agarwal, Sastry, Askell, Mishkin, Clark et~al.}]{radford2021learning}
Alec Radford, Jong~Wook Kim, Chris Hallacy, Aditya Ramesh, Gabriel Goh, Sandhini Agarwal, Girish Sastry, Amanda Askell, Pamela Mishkin, Jack Clark, et~al. 2021.
\newblock Learning transferable visual models from natural language supervision.
\newblock In \emph{International conference on machine learning}, pages 8748--8763.

\bibitem[{Robertson et~al.(1995)Robertson, Walker, Jones, Hancock-Beaulieu, Gatford et~al.}]{robertson1995okapi}
Stephen~E Robertson, Steve Walker, Susan Jones, Micheline~M Hancock-Beaulieu, Mike Gatford, et~al. 1995.
\newblock Okapi at trec-3.
\newblock \emph{Nist Special Publication Sp}, 109:109.

\bibitem[{Saito et~al.(2023)Saito, Sohn, Zhang, Li, Lee, Saenko, and Pfister}]{Saito2023Pic2Word}
Kuniaki Saito, Kihyuk Sohn, Xiang Zhang, Chun-Liang Li, Chen-Yu Lee, Kate Saenko, and Tomas Pfister. 2023.
\newblock Pic2word: Mapping pictures to words for zero-shot composed image retrieval.
\newblock In \emph{Proceedings of the IEEE Conference on Computer Vision and Pattern Recognition}.

\bibitem[{Sharma et~al.(2018)Sharma, Ding, Goodman, and Soricut}]{sharma2018conceptual}
Piyush Sharma, Nan Ding, Sebastian Goodman, and Radu Soricut. 2018.
\newblock Conceptual captions: A cleaned, hypernymed, image alt-text dataset for automatic image captioning.
\newblock In \emph{Proceedings of the 56th Annual Meeting of the Association for Computational Linguistics (Volume 1: Long Papers)}, pages 2556--2565.

\bibitem[{Su et~al.(2023)Su, Shi, Kasai, Wang, Hu, Ostendorf, Yih, Smith, Zettlemoyer, and Yu}]{su-etal-2023-one}
Hongjin Su, Weijia Shi, Jungo Kasai, Yizhong Wang, Yushi Hu, Mari Ostendorf, Wen-tau Yih, Noah~A. Smith, Luke Zettlemoyer, and Tao Yu. 2023.
\newblock \href {https://doi.org/10.18653/v1/2023.findings-acl.71} {One embedder, any task: Instruction-finetuned text embeddings}.
\newblock In \emph{Findings of the Association for Computational Linguistics: ACL 2023}, pages 1102--1121, Toronto, Canada. Association for Computational Linguistics.

\bibitem[{Vaze et~al.(2023)Vaze, Carion, and Misra}]{Vaze2023GeneCIS}
Sagar Vaze, Nicolas Carion, and Ishan Misra. 2023.
\newblock Genecis: A benchmark for general conditional image similarity.
\newblock In \emph{Proceedings of the IEEE Conference on Computer Vision and Pattern Recognition}.

\bibitem[{Wei et~al.(2024)Wei, Chen, Chen, Hu, Zhang, Fu, Ritter, and Chen}]{wei2024uniir}
Cong Wei, Yang Chen, Haonan Chen, Hexiang Hu, Ge~Zhang, Jie Fu, Alan Ritter, and Wenhu Chen. 2024.
\newblock Uniir: Training and benchmarking universal multimodal information retrievers.
\newblock In \emph{European Conference on Computer Vision}, volume 15145 of \emph{Lecture Notes in Computer Science}, pages 387--404. Springer.

\bibitem[{Wu et~al.(2021)Wu, Gao, Guo, Al-Halah, Rennie, Grauman, and Feris}]{wu2021fashioniq}
Hui Wu, Yupeng Gao, Xiaoxiao Guo, Ziad Al-Halah, Steven Rennie, Kristen Grauman, and Rogerio Feris. 2021.
\newblock Fashion iq: A new dataset towards retrieving images by natural language feedback.
\newblock In \emph{Proceedings of the IEEE/CVF Conference on Computer Cision and Pattern Recognition}, pages 11307--11317.

\bibitem[{Wu et~al.(2022)Wu, Tao, Shen, Xu, Geng, and Jiang}]{wu-etal-2022-pcl}
Qiyu Wu, Chongyang Tao, Tao Shen, Can Xu, Xiubo Geng, and Daxin Jiang. 2022.
\newblock \href {https://doi.org/10.18653/v1/2022.emnlp-main.826} {{PCL}: Peer-contrastive learning with diverse augmentations for unsupervised sentence embeddings}.
\newblock In \emph{Proceedings of the 2022 Conference on Empirical Methods in Natural Language Processing}, pages 12052--12066, Abu Dhabi, United Arab Emirates. Association for Computational Linguistics.

\bibitem[{Zhang et~al.(2024{\natexlab{a}})Zhang, Luan, Hu, Lee, Qiao, Chen, Su, and Chang}]{zhang2024magiclens}
Kai Zhang, Yi~Luan, Hexiang Hu, Kenton Lee, Siyuan Qiao, Wenhu Chen, Yu~Su, and Ming-Wei Chang. 2024{\natexlab{a}}.
\newblock Magiclens: Self-supervised image retrieval with open-ended instructions.
\newblock In \emph{International Conference on Machine Learning}.

\bibitem[{Zhang et~al.(2024{\natexlab{b}})Zhang, Zhang, Xie, Li, Dai, Long, Xie, Zhang, Li, and Zhang}]{zhang2024gme}
Xin Zhang, Yanzhao Zhang, Wen Xie, Mingxin Li, Ziqi Dai, Dingkun Long, Pengjun Xie, Meishan Zhang, Wenjie Li, and Min Zhang. 2024{\natexlab{b}}.
\newblock Gme: Improving universal multimodal retrieval by multimodal llms.
\newblock \emph{arXiv preprint arXiv:2412.16855}.

\end{thebibliography}

\end{document}